\documentclass[10pt,twocolumn,letterpaper]{article}

\usepackage[pagenumbers]{wacv} 

\usepackage{graphicx}
\usepackage{amsmath}
\usepackage{amssymb}
\usepackage{booktabs}

\usepackage{siunitx}
\usepackage{multicol}
\usepackage{multirow}
\usepackage{subcaption}
\usepackage{soul}
\usepackage[normalem]{ulem}
\usepackage[usenames,dvipsnames]{xcolor}

\usepackage[accsupp]{axessibility}

\usepackage[pagebackref,breaklinks,colorlinks]{hyperref}

\usepackage[capitalize]{cleveref}
\crefname{section}{Sec.}{Secs.}
\Crefname{section}{Section}{Sections}
\Crefname{table}{Table}{Tables}
\crefname{table}{Tab.}{Tabs.}

\begin{document}

\title{ Rotation-Constrained Cross-View Feature Fusion\\for Multi-View Appearance-based Gaze Estimation }

\author{Yoichiro Hisadome, Tianyi Wu, Jiawei Qin, Yusuke Sugano\\
Institute of Industrial Science, The University of Tokyo\\
Komaba 4-6-1, Tokyo, Japan \\
{\tt\small \{hisadome, twu223, jqin, sugano\}@iis.u-tokyo.ac.jp}
}

\maketitle

\newif\ifdraft
\drafttrue
\draftfalse

\newcommand{\Tref}[1]{Table~\ref{#1}}
\newcommand{\Eref}[1]{Eq.~\ref{#1}}
\newcommand{\Fref}[1]{Fig.~\ref{#1}}
\newcommand{\Aref}[1]{Algorithm~\ref{#1}}
\newcommand{\Sref}[1]{Sec.~\ref{#1}}

\begin{abstract}
Appearance-based gaze estimation has been actively studied in recent years. However, its generalization performance for unseen head poses is still a significant limitation for existing methods. This work proposes a generalizable multi-view gaze estimation task and a cross-view feature fusion method to address this issue. In addition to paired images, our method takes the relative rotation matrix between two cameras as additional input. The proposed network learns to extract rotatable feature representation by using relative rotation as a constraint and adaptively fuses the rotatable features via stacked fusion modules. This simple yet efficient approach significantly improves generalization performance under unseen head poses without significantly increasing computational cost. The model can be trained with random combinations of cameras without fixing the positioning and can generalize to unseen camera pairs during inference. Through experiments using multiple datasets, we demonstrate the advantage of the proposed method over baseline methods, including state-of-the-art domain generalization approaches.
The code will be available at \url{https://github.com/ut-vision/Rot-MVGaze}.
\end{abstract}


\section{Introduction}
\label{cha:introduction}

In anticipation of various applications such as robotics and accessibility, gaze estimation techniques have long been the subject of active research~\cite{Zhang2021_gaze_survey,Feng_ETRA_2021_typing,Bace_2020_ETRA_pursuit,Abdradou_ETRA_2021_password}. 
Appearance-based methods take the machine learning approach to directly estimate 3D gaze directions from input eye or face images, and have shown great potential for robust real-world gaze estimation~\cite{ghosh2021_gaze_survey2,Cheng2021_gaze_survey3}.
One of the key challenges in appearance-based gaze estimation is its limited generalization performance to unseen conditions.
The performance of gaze estimation models is often affected by various factors, including individuality, illumination, and the distributions of gaze and head pose.
Although various datasets have been proposed~\cite{xgaze,mpii_gaze,kellnhofer_ICCV2019_gaze360,eyediap,Fischer_ECCV_2018_rtgene,gaze_capture}, creating a generic model that can handle arbitrary conditions is still not a trivial task.

\begin{figure}
    \center
    \includegraphics[width=0.95\linewidth]{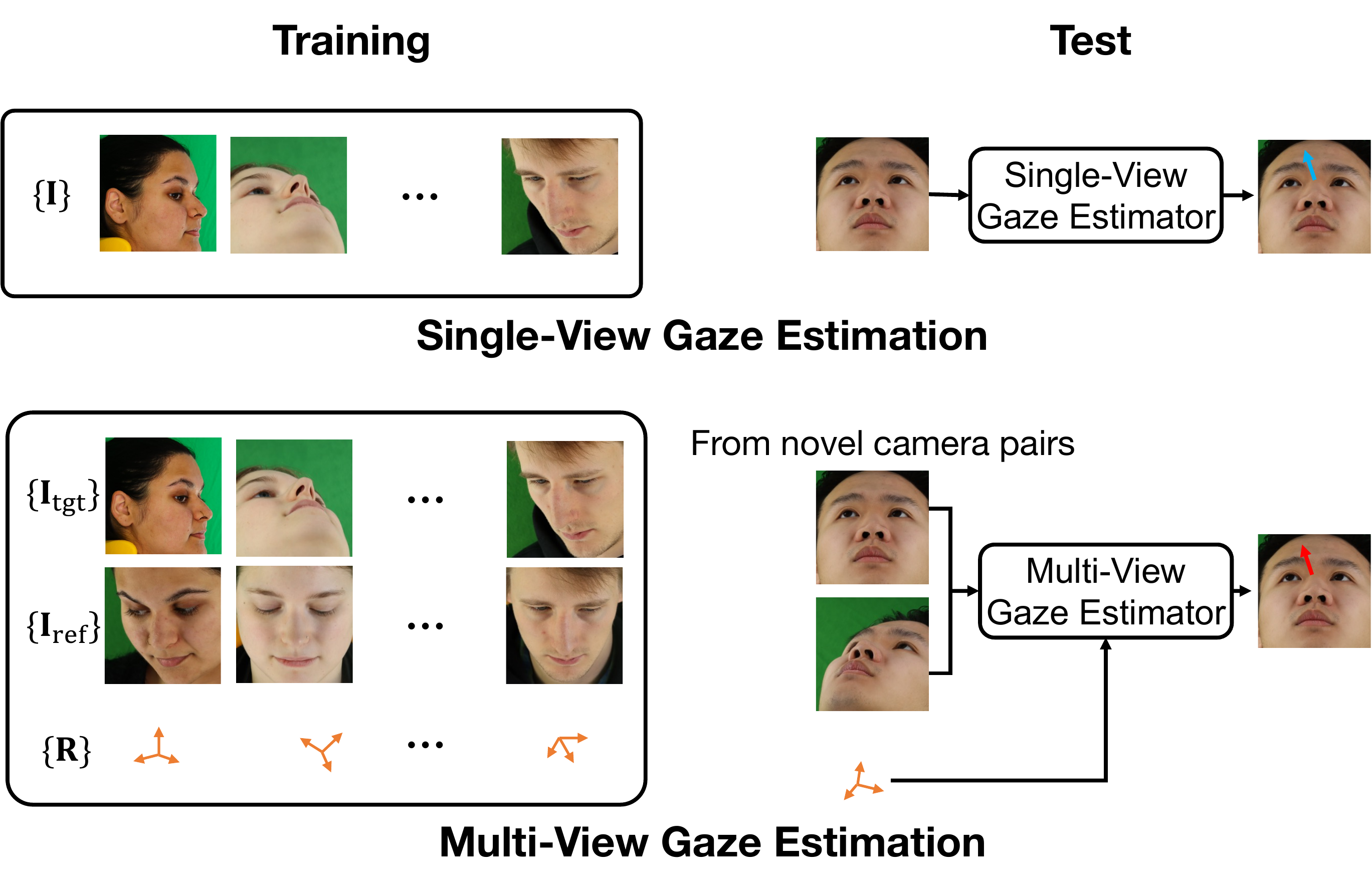}
    \caption{
        Overview of the proposed multi-view gaze estimation task.
        We estimate the 3D gaze direction from multiple synchronized images.
        The model can be generalized to unseen camera combinations unavailable during training by leveraging the relative rotation between cameras. 
    }
    \label{fig:teaser}
\end{figure}

Most of the existing appearance-based methods take monocular images as input and formulate gaze estimation as a task to estimate the gaze direction vector defined in the input image coordinate system.
For this reason, generalization difficulties in appearance-based gaze estimation vary greatly depending on the factors.
Specifically, unseen people and lighting conditions affect the facial appearance but do not fundamentally change the pattern of faces in the image.
In contrast, unseen head poses and their associated unseen gaze directions lead to entirely new patterns and have a more direct impact on the input-output relationship.
Therefore, it is usually more difficult for existing gaze estimation models to generalize to unseen head poses.

If not limited to appearance-based methods, multi-camera geometry-based eye tracking systems have long been the subject of active research~\cite{ ruddarraju2003perceptual, ohno2004free, shih2004novel, utsumi2012gaze, arar2015robust}. 
Such a multi-view approach may solve the above problems in appearance-based methods.
For many application scenarios, such as driver monitoring and public displays, using multiple synchronized machine vision cameras is a sufficiently realistic assumption for appearance-based gaze estimation.
The expected effect of multi-view input is not limited to simply increasing information and improving accuracy.
The model could acquire a head pose-independent feature representation by training a gaze estimation model considering the geometric positional relationship between input images.
However, considering the cost of training data acquisition, training a model specialized for cameras in a particular positional relationship is not practical.
The key challenge is to train a model that can perform accurate gaze estimation even if the camera's positional relationship changes between inference and training.

This work proposes a multi-view appearance-based gaze estimation method that utilizes the relative rotation between cameras as additional input information (\Fref{fig:teaser}). 
Assuming the normalization process used in appearance-based gaze estimation~\cite{normalize}, a relative rotation matrix can always express the interrelationship of camera positions. 
The main idea of the proposed method is to use the rotation matrix as a constraint for feature fusion between images.
The proposed method consists of stacked rotation-constrained feature fusion blocks that can be combined with arbitrary feature extraction backbones.
In each block, one of the features is multiplied by the rotation matrix to transfer to the other image.
Although the physical rotation is not originally applicable to the feature space, the model is expected to learn to extract rotatable features through the explicit training process incorporating the rotation operation.
We demonstrate that our method acquires rotatable feature representation through experimental analyses on multiple datasets~\cite{xgaze,jiawei}.
The proposed method achieves better generalizability than baseline approaches, including state-of-the-art domain generalization methods.

Our key contributions are threefold.
First, this paper addresses the camera-independent multi-view appearance-based gaze estimation task for the first time in the literature.
Second, we propose a novel cross-view feature fusion approach incorporating the relative rotation matrix into multi-view gaze estimation.
Our method uses the rotation matrix as a constraint to transfer features between images, and we provide a thorough analysis of the internal feature representation.
Third, we demonstrate that multi-view gaze estimation improves generalization performance for unseen head poses.
Through experiments, we show that the accuracy gains from multi-view training are superior to state-of-the-art methods.
We also provide thorough analyses and visualizations of the internal feature representation obtained through the rotation constraint.


\section{Related Work}

\paragraph{Appearance-based Gaze Estimation.}
Appearance-based gaze estimation is a task to regress 3D gaze directions from full-face~\cite{xgaze,puregaze,cheng_ICPR2022_gazetr,normalize,full-face,jiawei,eyetracking-everyone} or eye-region images~\cite{deep-pict,in-wild,sugano_cvpr2014_learning-by-synthesis,Yu_CVPR2020, mpii_gaze,eyediap, cheng_TIP2020_twoeyeasymmery,cheng_ECCV2018_evaluation-guided}.
In recent years, the advances in deep neural networks have enabled gaze estimation techniques with decent within-dataset performance~\cite{Chen_2019_ACCV_CANet,full-face,xgaze,cheng_ICPR2022_gazetr,Cheng_2020_AAAI_coarse_to_fine}.
However, noticeable performance degradation can be observed when deployed in real-world applications.
Consequently, the performance of state-of-the-art approaches would be limited under unconstrained conditions, primarily when the gap arising from the aforementioned factors is significantly large during training and testing.
While there are ongoing attempts on personalization~\cite{few_shot_yu2019improving, faze}, and domain adaptation~\cite{gaze_rot_consistent, liu_ICCV2021_PnP_GA, simgan_2017_CVPR}, training models that can generalize to unknown environments is a significant challenge.
This study addresses this domain gap issue by improving the generalization ability to unseen head poses.

Given the various factors to consider for generalizing appearance-based gaze estimation, previous studies have proposed many datasets.
Because it is difficult to cover all factors concurrently, many datasets were constructed focusing on a certain diversity~\cite{gaze_capture,mpii_gaze,eyediap}.
Covering a variety of head poses is challenging, and each recent dataset still has limitations, such as the lighting conditions diversity~\cite{xgaze} and the gaze labels accuracy~\cite{kellnhofer_ICCV2019_gaze360}.
Although intended for a monocular estimation task, the ETH-XGaze dataset~\cite{xgaze} was created using multiple synchronized cameras and can be used for multi-view purposes.
Recent work on synthesizing face images with ground-truth gaze directions further indicates the possibility of acquiring training data for multi-view estimation~\cite{jiawei, sted, ruzzi2022gazenerf, yin2022nerf}.
This study examines the novel task of multi-view gaze estimation on these datasets.

\paragraph{Domain Generalization for Gaze Estimation.} 
There are some prior attempts to alleviate the gap between training and testing environments through domain generalization.
Some prior work~\cite{farkhondeh2022towards, kothari2021weakly_laeo} use large-scale unlabeled face images for pretraining or as an additional training signal to generalize gaze estimator.
However, these approaches still require extra samples from either the target domain or the Internet, which is often nontrivial to be prepared in practice even without ground-truth gaze labels.
The proposed method differs from these approaches because it does not use extra data for achieving generalization.

Another line of work on domain generalization directly improves model robustness on unseen domains by removing person-dependent factors during training and is thus closer to our objective~\cite{faze,puregaze}.
We note that the goal of multi-view gaze estimation is not only domain generalization, and the direction of this work is not strictly consistent with them.
Nevertheless, our approach improves robustness on unseen head poses, which none of the above-mentioned methods have explicitly proven.

\paragraph{Multi-view Feature Fusion.}
Most previous works on multi-view eye tracking take the model-based approach~\cite{ruddarraju2003perceptual, ohno2004free, shih2004novel, arar2015robust, arar2017robust}, which requires a more complex setup with external light sources.
Although some image-only multi-view methods exist~\cite{utsumi2012gaze}, they still rely on geometric eyeball models.
Prior research has shown that such geometry or shape-based approaches are inferior to appearance-based methods in terms of performance~\cite{zhang2019evaluation}.
In contrast, appearance-based multi-view gaze estimation~\cite{gideon2022unsupervised,fix_stereo} has been understudied.
Lian~\etal~\cite{fix_stereo} proposed directly concatenating features from stereo images to predict 2D on-screen gaze positions.
Gideon~\etal~\cite{gideon2022unsupervised} proposed disentangling image features through feature swapping between multi-view videos, different from our frame-by-frame setting.
However, these methods require fixed cameras during training and testing, and their effectiveness in unknown camera configurations remains unproven.
The proposed method uses the relative rotation matrix between camera pairs as additional input to overcome this drawback, achieving multi-view gaze estimation generalizable to unseen camera pairs.

Multi-view input has also been explored in many computer vision tasks, but the direct use of these methods for gaze estimation is not straightforward. 
Recent work on multi-view stereo constructs and uses 3D voxel representation from multi-view features~\cite{mvsnet, pyramid-mvs, cascade-cost,epps}.
While such a voxel representation is suited to geometry-related tasks, it is not directly applicable to gaze estimation, which is rather an attribute regression task.
NeRF~\cite{nerf, neural_rendering, nerface, nerfwild, neuralvoxel, keypointnerf} can also be applied to gaze redirection and training data synthesis~\cite{yin2022nerf,ruzzi2022gazenerf}, but it is not yet directly related to estimation tasks.
Unlike these approaches strongly based on physical geometry, our method uses relative camera rotation as a soft constraint for learnable feature extraction and fusion blocks.


\section{Method}

\begin{figure*}[t]
    \centering
    \includegraphics[width=0.9\linewidth]{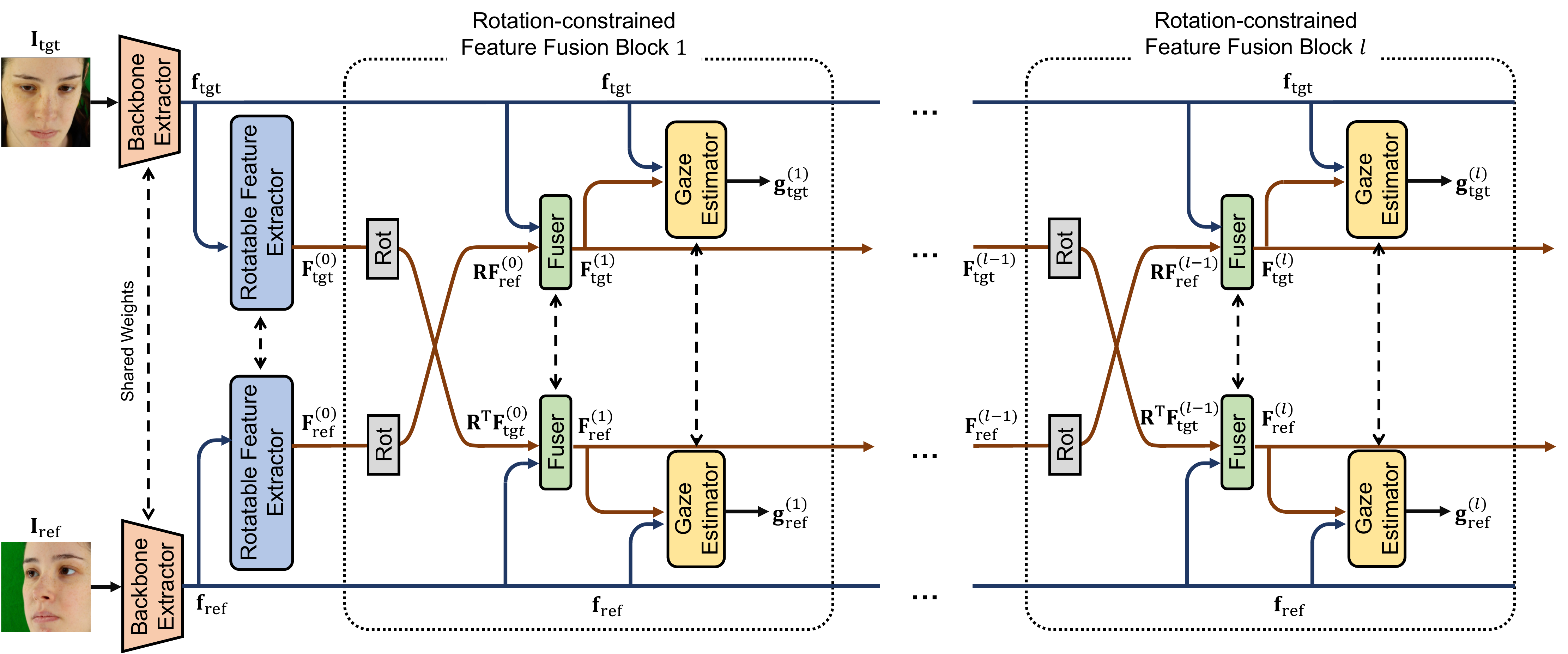}
    \caption{
        The overview of the proposed network which consists of stacked rotation-constrained feature fusion blocks.
        The subscripts \emph{ref} and \emph{tgt} indicate reference and target images, respectively, and the superscripts denote block id.
        While network components share weights between the target and reference sides, they do not share weights across stacked blocks.
    }
    \label{fig:iter_rotate}
\end{figure*}

This work aims to design a network to efficiently perform feature transitions and fusions between images according to the input rotation matrix.
Since rotation has an extremely low dimensionality compared to image features, it is not optimal to feed it into the network as another feature.
The proposed method achieves this goal via stacked rotation-constrained feature fusion blocks.

\subsection{Overview}

\Fref{fig:iter_rotate} shows the overview of the proposed network, which consists of stacked rotation-constrained feature fusion blocks.
The inputs are the target image $\mathbf{I}_\mathrm{tgt}$ and reference image $\mathbf{I}_\mathrm{ref}$.
As described earlier, these face images are assumed to be normalized for appearance-based gaze estimation~\cite{normalize}, and their mutual relationship can be fully described by the rotation matrix $\mathbf{R}$. 
$\mathbf{R}$ indicates the rotation from the reference camera to the target camera coordinate systems and can be obtained either from the extrinsic camera calibration, or head poses estimated through the normalization process.
While the goal is to estimate the gaze direction $\mathbf{g}_{\mathrm{tgt}}$ of the target image $\mathbf{I}_\mathrm{tgt}$, the role of the target $\mathbf{I}_\mathrm{tgt}$ and the reference $\mathbf{I}_\mathrm{ref}$ are symmetrical in our method.
The network components on both sides thus share weights, but the stacked fusion blocks do not share weights.

Given input images, the proposed method first extracts two features.
The network first extracts the backbone feature vectors $\mathbf{f}_{\mathrm{tgt}}$ and $\mathbf{f}_{\mathrm{ref}}$ using the \emph{Backbone Extractor} module.
The \emph{Rotatable Feature Extractor} module then takes these backbone features as inputs and outputs $D$ three-dimensional vectors.
These vectors are stacked to form rotatable feature tensors $\mathbf{F}^{(0)}_{\mathrm{ref}}, \mathbf{F}^{(0)}_{\mathrm{tgt}} \in \mathbb{R}^{3 \times D}$.
While backbone features are used unaltered throughout the process, the rotatable features are updated through the rotation-constrained feature fusion blocks.
The output gaze vectors $\mathbf{g}_{\mathrm{tgt}}$ and $\mathbf{g}_{\mathrm{ref}}$ are defined as 3D unit vectors in each normalized camera coordinate system.

\subsection{Rotation-Constrained Feature Fusion}
\label{subsec:iter_rotate}

As discussed earlier, the basic idea behind the rotation-constrained feature fusion block is directly applying the rotation (multiplying the rotation matrix) in the feature space.
We expect the network to learn to extract rotatable feature representation by introducing a rotation-constrained feature fusion mechanism.
The rotated and transferred features are expected to complement the information in each image, and therefore the optimal features cannot be obtained simply by observing individual images.
The proposed method is designed to achieve optimal feature transfer by repeating the process of fusing the rotated features with the backbone features of the destination.

In the $i$-th block, the model first multiplies the rotation matrix $\mathbf{R}$ to the reference rotatable feature $\mathbf{F}^{(i-1)}_{\mathrm{ref}}$.
The \emph{Fuser} module then fuses the rotated feature with the backbone feature $\mathbf{f}_\mathrm{tgt}$ of the target image to obtain an updated rotatable feature $\mathbf{F}^{(i)}_\mathrm{tgt}$ of the target image. 
The network is symmetric and applies the same operation to update the rotatable feature of the reference image using the back-rotated reference feature $\mathbf{R}^{\top} \mathbf{F}^{(i)}_\mathrm{tgt}$.
The \emph{Gaze Estimator} modules then output intermediate estimates of gaze directions $\mathbf{g}^{(i)}_\mathrm{tgt}$ and $\mathbf{g}^{(i)}_\mathrm{ref}$ using both backbone and rotatable features.
The model repeats the above fusion process until $l$ blocks, and the $l$-th estimated gaze $\mathbf{g}^{(l)}_{\mathrm{tgt}}$ becomes the final output.

The loss function $\mathcal{L}_\mathrm{total}$ is defined for all intermediate outputs $\mathbf{g}^{(i)}$, not just the final output $\mathbf{g}^{(l)}$ as
\begin{equation}
\mathcal{L}_\mathrm{total} = \sum^l_{i=1}{\alpha^{l - i} \cdot \mathcal{L}_{i}} (\mathbf{g}_{\mathrm{tgt}}^{(i)}, \mathbf{g}_{\mathrm{ref}}^{(i)}).
\end{equation}
$\mathcal{L}_{i}$ is the angular loss for the $i$-th block defined as
\begin{equation}
\mathcal{L}_{i} = \arccos{(\mathbf{g}_{\mathrm{tgt}}^{(i)\top}\hat{\mathbf{g}}_{\mathrm{tgt}})} + \arccos{(\mathbf{g}_{\mathrm{ref}}^{(i)\top}\hat{\mathbf{g}}_{\mathrm{ref}})},
\end{equation}
where $\hat{\mathbf{g}}_{\mathrm{tgt}}$ and $\hat{\mathbf{g}}_{\mathrm{ref}}$ indicate the ground-truth 3D gaze directions corresponding to target and reference images.
$\alpha$ is a hyperparameter indicating the decay to gaze estimated from earlier blocks.

\subsection{Rotation Matrix}

As discussed earlier, the relative rotation matrix $\mathbf{R}$ can be obtained either from camera calibration or head pose estimation.
$\mathbf{R}$ describes the rotation term between the normalized cameras.
The meaning of $\mathbf{R}$ is the same in either approach, and the results are the same if there is no head pose estimation error.
Although a translation vector $\mathbf{t}$ is also needed to describe the relationship between two cameras completely, it is uniquely determined by $\mathbf{R}$ and can be ignored under the assumption of data normalization~\cite{normalize}.

For the first option based on camera calibration, the rotation matrix can be calculated using the camera extrinsic parameters and the normalization matrices~\cite{normalize} as $\mathbf{R} = \mathbf{N}_{\mathrm{tgt}} \tilde{\mathbf{R}} \mathbf{N}^{\top}_{\mathrm{ref}}$.
$\tilde{\mathbf{R}}$ is the rotation matrix obtained via camera calibration and therefore corresponds to the coordinate systems of the original camera before normalization.
$\mathbf{N}_{\mathrm{tgt}}$ and $\mathbf{N}_{\mathrm{ref}}$ are the normalization matrices, which indicate the transformation from the original to the normalized camera coordinate systems.
For the second option, the rotation is calculated using head poses estimated through the normalization process.
If we denote the rotation from the head coordinate system to the normalized camera coordinate system as $\mathbf{H}_\mathrm{tgt}$ and $\mathbf{H}_\mathrm{ref}$, the rotation matrix is $\mathbf{R} = \mathbf{H}_\mathrm{tgt} \mathbf{H}^{\top}_\mathrm{ref}$.

\subsection{Implementation Details}

Unless otherwise noted, we set the number of blocks $l=3$ in all experiments.
We used ResNet-50~\cite{resnet} as the \emph{Backbone Extractor} module, which was initialized with pre-trained weights on ImageNet~\cite{imagenet} and fine-tuned through the training process.
We used two-layered MLPs for the \emph{Rotatable Feature Extractor} and the \emph{Gaze Estimator}, and a three-layered MLP for the \emph{Fuser}, with $D=512$.
The \emph{Rotatable Feature Extractor} receives the backbone feature vector from the \emph{Backbone Extractor} and outputs three $D$-dimensional vectors.
These feature vectors are stacked to form a $3 \times D$ rotatable feature matrices.
The \emph{Fuser} and \emph{Gaze Estimator} first flatten the $3\times D$ shaped rotatable feature matrices.
The flattened matrices are then concatenated with the backbone feature vectors from the other view.
Like the \emph{Rotatable Feature Extractor}, the \emph{Fuser} also outputs three $D$-dimensional vectors stacked to form a $3 \times D$ matrix.
All MLPs use ReLU~\cite{relu} as the activation layer.

During the training, we applied random mask data augmentation to the input images to force the model to exploit features from another view.
Inspired by random erasing~\cite{random_erasing}, we masked the input image with multiple randomly sized ($5$--$30$\% of the image width) small squares with a 50\% probability.
The quantity of squares was determined randomly so that the proportion of the total area covered by the squares was restricted to $50$--$60$\% of the face image. 
We also applied color jitter, translation, and scaling.
We set the saturation, brightness, and contrast range to $0.1$.
The intensities of translation and scaling were set to be small ($0.01$ for translation, $0.99$--$1.01$ for scaling) to represent possible face alignment errors during the normalization process.
We apply the same data augmentation to all baseline methods for fair comparisons.

We set the training batch size to $256$.
We used Adam~\cite{adam} optimizer with a weight decay of \num{1d-6}.
We used CyclicLR~\cite{cycliclr} as the learning rate scheduler, with the base and maximum learning rate of \num{1d-6} and \num{1d-3}, decaying 0.5 per cycle.
The cycle steps were determined so that one cycle was completed in one epoch.
We used mixed precision training and set the block decay $\alpha$ to 0.5.


\section{Experiments}
\label{sec:experiment}

\subsection{Experimental Setting}\label{subsec:setting}


We use two datasets that provide synchronized multiple views of participants and corresponding 3D gaze directions.
\textbf{ETH-XGaze}~\cite{xgaze} contains 110 participants, each captured with 18 cameras simultaneously.
Since one of our goals is to evaluate the generalization performance against unseen head poses, we split the training subset of 80 participants instead of using their official test data.
We directly use the camera extrinsic parameters and head pose estimation results provided with the dataset.
\textbf{MPII-NV} is a synthetic dataset created by following Qin~\etal's approach~\cite{jiawei}.
The dataset is based on the MPIIFaceGaze dataset~\cite{mpii_gaze} that consists of monocular images of 15 participants.
3D face meshes are first reconstructed from the original MPIIFaceGaze images and then rotated with their ground-truth gaze vector to generate face images of new head poses.
We synthesized the data so that the head pose distribution is the same as that of the ETH-XGaze training set.
Since it is a purely synthetic dataset, we can use the camera position for image rendering to compute the relative rotation matrix.
After data normalization, the input image resolution for both datasets is $224 \times 224$.
Both datasets were collected with IRB approval or consent from participants.

We used $k$-fold cross-validation regarding participant IDs ($k=4$ for ETH-XGaze and $k=3$ for MPII-NV), and all methods were trained for 15 epochs without validation data.
Given the practical scenario where the camera positions at the deployment are not necessarily known at the training time, generalization performance for unseen camera positions is an important metric.
Therefore, we further split the cameras into training and test sets in each fold to evaluate the generalization performance.
Specifically, we split the $18$ cameras of both ETH-XGaze and MPII-NV into $12$ for training and $6$ for testing, three of which are within the head-pose range of the training camera set (interpolation), and the other three are outside (extrapolation).
Please note that head poses with respect to the camera position are nearly fixed under the ETH-XGaze setup, and there is a strong correlation between the head poses and camera positions.
Training and testing image pairs are constructed by randomly selecting two cameras from each set.


\textbf{Concat} represents the approach of Lian~\etal~\cite{fix_stereo} as the baseline multi-view appearance-based method.
It extracts features from two images using weight-shared ResNet-50 and then estimates the gaze direction using concatenated features.
\textbf{Single} is the single-image baseline method corresponding to the one reported in ETH-XGaze paper~\cite{xgaze}.
It extracts features from the monocular input image using ResNet-50~\cite{resnet} followed by a fully-connected layer to output gaze direction. 
\textbf{Gaze-TR}~\cite{cheng_ICPR2022_gazetr} is one of the state-of-the-art methods for single-image gaze estimation. 
We adopted the hybrid version containing a ResNet-50~\cite{resnet} extractor and a transformer encoder~\cite{transformers, vit}.
It extracts features from ResNet-50~\cite{resnet} and feeds the feature maps to the transformer encoder, followed by an MLP to output the gaze directions.
\textbf{Frontal Selection} is fundamentally a single-view model, but multi-view information is utilized naively during inference.
It predicts the gaze based on the more frontal image from the reference and target images.

Since our goal is generalizable gaze estimation, we also include some single-image domain generalization approaches as baselines.
Since our method requires no prior knowledge of the target domain, we excluded domain adaptation methods which require target domain data.
Please note that unsupervised domain adaptation still requires target domain images.
\textbf{PureGaze}~\cite{puregaze} introduced an extra CNN-based image reconstruction module to the ResNet-18 backbone and MLP.
We followed the official implementation for MLP and the reconstruction module while replacing the backbone with ResNet-50.
\textbf{DT-ED}~\cite{faze} first extracts the latent codes of appearance, gaze, and head pose from the source image, then a decoder is used to reconstruct the target image from the rotated head pose and gaze features, and an MLP is used to predict gaze directions from the gaze features only.
We follow the original structure of 4-block DenseNet~\cite{densenet} and a growth rate of 32, and the target image was randomly chosen from the same subject.

\subsection{Performance Comparison}\label{sec:baseline}

\begin{table}[t]
\begin{subtable}{\linewidth}
    \centering
    \scalebox{0.9}{
        \begin{tabular}{@{}lcccc@{}} \toprule
            Train/Test      & \multicolumn{2}{c}{XGaze (Calib.)} & \multicolumn{2}{c}{MPII-NV} \\
            Head pose                              & Seen            & Unseen          & Seen            & Unseen \\ \midrule
            Single                                 & 4.07\textdegree & 6.97\textdegree & 7.45\textdegree & 8.60\textdegree \\
            Gaze-TR~\cite{cheng_ICPR2022_gazetr} & 4.16\textdegree & 7.24\textdegree & 7.38\textdegree & 8.47\textdegree \\
            DT-ED~\cite{faze}                       & 5.07\textdegree & 7.88\textdegree & 7.91\textdegree & 9.44\textdegree \\
            PureGaze~\cite{puregaze}               & 3.99\textdegree & 7.98\textdegree & 7.59\textdegree & 9.13\textdegree \\
            \midrule
            Frontal Selection                 & 3.70\textdegree & 5.58\textdegree & 7.04\textdegree & 7.40\textdegree \\
            Concat~\cite{fix_stereo}               & 3.71\textdegree & 8.60\textdegree & \textbf{6.78\textdegree} & 8.41\textdegree \\
            Proposed                               & \textbf{3.50\textdegree} & \textbf{4.95\textdegree} & 6.81\textdegree & \textbf{7.06\textdegree} \\ \bottomrule
        \end{tabular}
    }
    \caption{
        Within-dataset evaluation.
    }
    \label{tab:within-calib}
\end{subtable}

\begin{subtable}{\linewidth}
    \centering
    \scalebox{0.9}{
        \begin{tabular}{@{}lcccc@{}} \toprule
            Train  & \multicolumn{2}{c}{MPII-NV} & \multicolumn{2}{c}{XGaze (Calib.)} \\
            Test      & \multicolumn{2}{c}{XGaze (Calib.)} & \multicolumn{2}{c}{MPII-NV} \\
            Head pose                              & Seen            & Unseen          & Seen            & Unseen \\ \midrule
            Single                                 & 18.44\textdegree & 18.40\textdegree & 19.71\textdegree & 18.10\textdegree \\
            Gaze-TR~\cite{cheng_ICPR2022_gazetr}   & 18.90\textdegree & 17.35\textdegree & 15.34\textdegree & 17.06\textdegree \\
            DT-ED~\cite{faze}                      & 14.39\textdegree & 17.20\textdegree &  24.08\textdegree &  27.05\textdegree \\
            PureGaze~\cite{puregaze}               & 18.63\textdegree & 22.59\textdegree & 14.52\textdegree & 14.86\textdegree \\
            \midrule
            Frontal Selection                      & 19.23\textdegree & 19.23\textdegree & 16.16\textdegree & 14.71\textdegree \\
            Concat~\cite{fix_stereo}               & 17.73\textdegree & 17.82\textdegree & 14.43\textdegree & 14.23\textdegree \\
            Proposed                               & \textbf{14.27\textdegree} & \textbf{14.55\textdegree} & \textbf{12.60\textdegree} & \textbf{12.07\textdegree} \\ \bottomrule
        \end{tabular}
    }
    \caption{
            Cross-dataset evaluation.
    }
    \label{tab:cross-eval}
\end{subtable}
    \caption{
            Within- and cross-dataset evaluation using rotation from camera calibration.
            Each number indicates the mean angular error of the proposed and baseline methods.
    }
    \label{tab:withi-cross-eval}
\end{table}

\paragraph{Within-Dataset Evaluation.}

We first conduct a within-dataset evaluation on ETH-XGaze and MPII-NV. 
For both datasets, we present the cases where the relative rotation matrix $\mathbf{R}$ is from the camera extrinsic calibration.
We report the angular error averaged over the $k$ folds in \Tref{tab:within-calib}.
The \emph{Head pose} column corresponds to the split of the cameras as described in \Sref{subsec:setting}. 
For ETH-XGaze under the seen head pose condition, multi-view approaches (\emph{Concat} and \emph{Proposed}) consistently outperform single-image methods.
However, the \emph{Concat} model shows lower accuracy in the unseen head pose condition than the single-image baselines.
This shows the difficulty of learning a generic head pose-independent gaze feature under the multi-view condition.
Our proposed method with feature rotation and repetitive fusion achieves the best performance.
In particular, it improves $0.57$\textdegree~($14.0\%$) over the \emph{Single} baseline in the seen head pose condition, and more prominently, improves $1.54$\textdegree~($29.0\%$) in the unseen head pose condition. 
This demonstrated the significance of our proposed method for novel-view-generalizable gaze estimation.

For MPII-NV, the proposed method outperforms all single-image and multi-view baselines under unseen head pose conditions.
The performance improvement is $8.6\%$ and $17.9\%$ over the \emph{Single} baseline in the seen and unseen head pose conditions, respectively. 
It is also worth noting that while \emph{Concat} and our proposed method perform almost equivalently well in the \emph{seen} condition, our proposed method outperforms \emph{Concat} by $16.1\%$ in the \emph{unseen} condition.
This proves the advantage of the proposed feature rotation and repetitive feature fusion in fusing head-pose-independent representations.

\begin{figure}[t]
    \centering
    \includegraphics[width=0.9\linewidth]{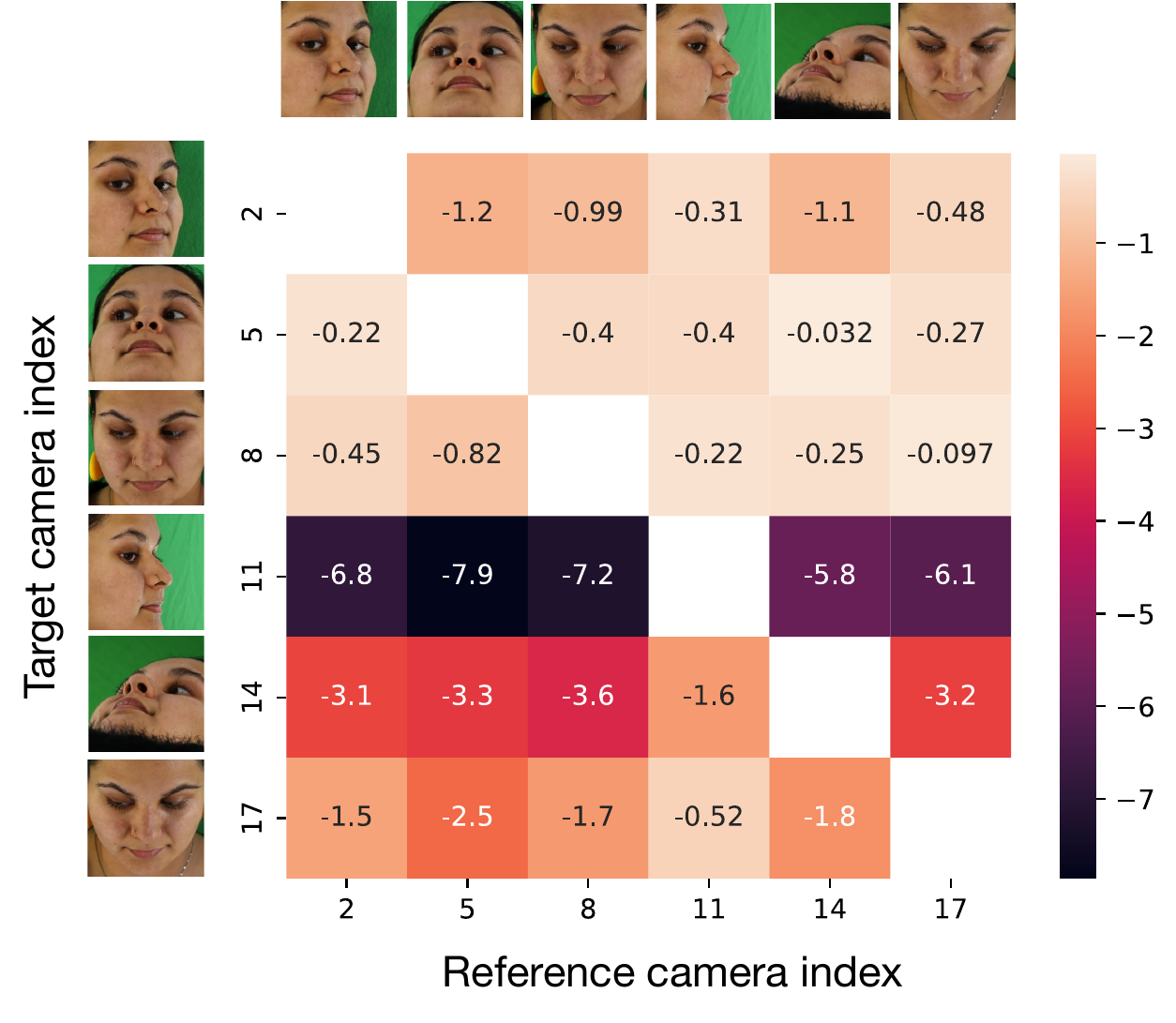}
    \caption{
        Visualization of the performance gain from multi-view input.
        The numbers on the x and y axis indicate the camera index in ETH-XGaze.
        The numbers and colors in the matrix indicate the mean gaze error between the gaze estimation errors of the proposed method and \emph{Single} model.
    }
    \label{fig:single-vs-proposed}
\end{figure}

\Fref{fig:single-vs-proposed} further visualizes the difference of mean gaze error between the proposed and \emph{Single} model in the ETH-XGaze unseen head pose condition.
We can see that the gaze estimation errors drastically decrease when the target head pose corresponds to extrapolation (cameras 11, 14, 17) and the reference head pose corresponds to interpolation (cameras 2, 5, 8).
We can also confirm a decent error reduction even when the head poses correspond to extrapolation.

\paragraph{Cross-Dataset Evaluation.}

We further evaluate the performance of the proposed method in the cross-dataset setting.
We train on one of ETH-XGaze and MPII-NV and evaluate angular errors on the other dataset.
Since the two datasets contain different participants, we use all participants in one dataset for training and all participants in the other for the test.
\Tref{tab:cross-eval} shows the results of the cross-dataset evaluation in the unseen and seen head pose conditions.
We can observe the same tendencies as within-dataset evaluation in \Tref{tab:within-calib}.
In both conditions, the accuracy of our proposed method is much higher than any of the baseline methods.
From these results, it can be seen that the proposed method is also effective in reducing the inter-domain gap.
The proposed method performs better than the \emph{Concat} baseline, indicating that the reduction of the inter-domain gap owes to the rotation-constrained feature fusion rather than multi-view estimation.


\begin{table}[t]
    \centering
    \scalebox{0.9}{
        \begin{tabular}{@{}lrrc@{}}
            \toprule
             Head pose                      & Seen                      & Unseen                    & Infer. time   \\ \midrule
             w/o Rotation matrix            & 3.64\textdegree           & 9.00\textdegree           &      -         \\
             MLP encoding                   & 3.66\textdegree           & 8.22\textdegree           &      -        \\ \bottomrule
             \# block = 1                   & \textbf{3.49\textdegree}  & 5.22\textdegree           & 22.9 ms       \\
             \# block = 2                   & 3.50\textdegree           & 4.99\textdegree           & 23.0 ms       \\
             \# block = 3 (Proposed)   & 3.50\textdegree           & \textbf{4.95\textdegree}  & 24.8 ms       \\
             \# block = 4                   & 3.50\textdegree           & 4.97\textdegree           & 25.2 ms      \\ \bottomrule
        \end{tabular}
    }
    \caption{Ablation studies of the rotation encoding approaches and the number of fusion blocks. Inference time is benchmarked on a single NVIDIA V100 GPU.}
    \label{tab:ablation-rot-encoding}
\end{table}

\begin{table}[t]
    \centering
    \scalebox{0.9}{
    \begin{tabular}{@{}lcc@{}} \toprule
         Test  & \multicolumn{2}{c}{XGaze (Pose)}  \\
         Head pose  & Seen & Unseen \\ \midrule
         Concat~\cite{fix_stereo}  & 3.71\textdegree & 8.60\textdegree \\
         Proposed (Calib.)  &      4.63\textdegree & \textbf{5.68\textdegree} \\
         Proposed (Pose.)   & \textbf{3.63\textdegree} & 7.12\textdegree \\ \bottomrule
    \end{tabular}
    }
    \caption{
        ETH-XGaze within-dataset evaluation using rotation matrices obtained from head pose without calibration.
        Each number indicates the mean angular error.
    }
    \label{tab:from_headpose}
\end{table}

\subsection{Detailed Performance Analyses}\label{sec:rotation-analysis}

\paragraph{Ablation Studies.}

In \Tref{tab:ablation-rot-encoding}, we compare different usage of the rotation matrix.
The second row (\emph{w/o Rotation matrix} corresponds to the model that uses the stacked architecture but concatenates the features without rotation.
The third row (\emph{MLP Encoding}) is the case where the rotation matrix is used as an additional feature instead of multiplication at each block.
In this case, we concatenated the flattened rotation matrices with the features and then fed them to an MLP encoder which is almost the same as \emph{Fuser} except for the input feature dimension.
As with the proposed model, the learnable weights are shared for target and reference but not shared across blocks for both cases.
Although feeding flattened rotation matrices (\emph{MLP Encoding}) improves the accuracy from the \emph{Single}, it is still inferior to the proposed method.
We also change the number of fusion blocks from the fourth to the last row in \Tref{tab:ablation-rot-encoding}.
The impact of the stacked fusion blocks is different under unseen head pose conditions.
However, the three-block model has the best performance for unseen head poses and is almost the best for the seen head pose condition.

The rightmost column in \Tref{tab:ablation-rot-encoding} shows the inference times of our method on NVidia V100.
\emph{Single} model's inference time is 10.7 ms and \emph{Gaze-TR} is 35.0 ms.
We can observe that the additional inference cost from additional fusion blocks is relatively minor, and the inference time is almost double that of the \emph{Single} baseline.
Although the increase in computational cost is one limitation of the proposed method, it is still faster than more complex single-image methods such as \emph{Gaze-TR} and is considered to be well within the practical range.

\paragraph{Accuracy of Rotation Matrix.}

In \Tref{tab:from_headpose}, we compare the performance of the proposed method using a rotation matrix obtained from estimated head poses without camera calibration.
Since the head pose is expected to be perfectly accurate on synthetic MPII-NV, we only evaluate the cases using real images from ETH-XGaze.
\emph{Proposed (Calib.)} and \emph{Proposed (Pose)} correspond to the cases where the rotation matrices for training data are obtained from calibration and head pose, respectively.
As a reference, we also show the performance of the \emph{Concat} model.
Please note that all baseline methods, including \emph{Concat}, do not use a rotation matrix as input. 
Thus, the numbers are the same as \Tref{tab:within-calib}.

It can be seen that the proposed model trained with calibration is sensitive to the noise of the rotation matrix at inference times.
However, \emph{Proposed (Calib.)} method still performs best for the unseen head pose condition.
If the model is trained with rotation matrices from head pose (\emph{Proposed (Pose)}), unseen head pose performance degrades while the performance is improved for the seen head pose condition.

While the \emph{Frontal Selection} in Table~\ref{tab:withi-cross-eval} shows that simply utilizing multi-view information can improve performance from the \emph{Single}, our proposed approach performs best in most cases, demonstrating the significance of our rotation-constrained feature fusion.

\subsection{Rotatable Feature Representation}

\begin{figure}[t]
    \centering
    \includegraphics[width=0.9\linewidth]{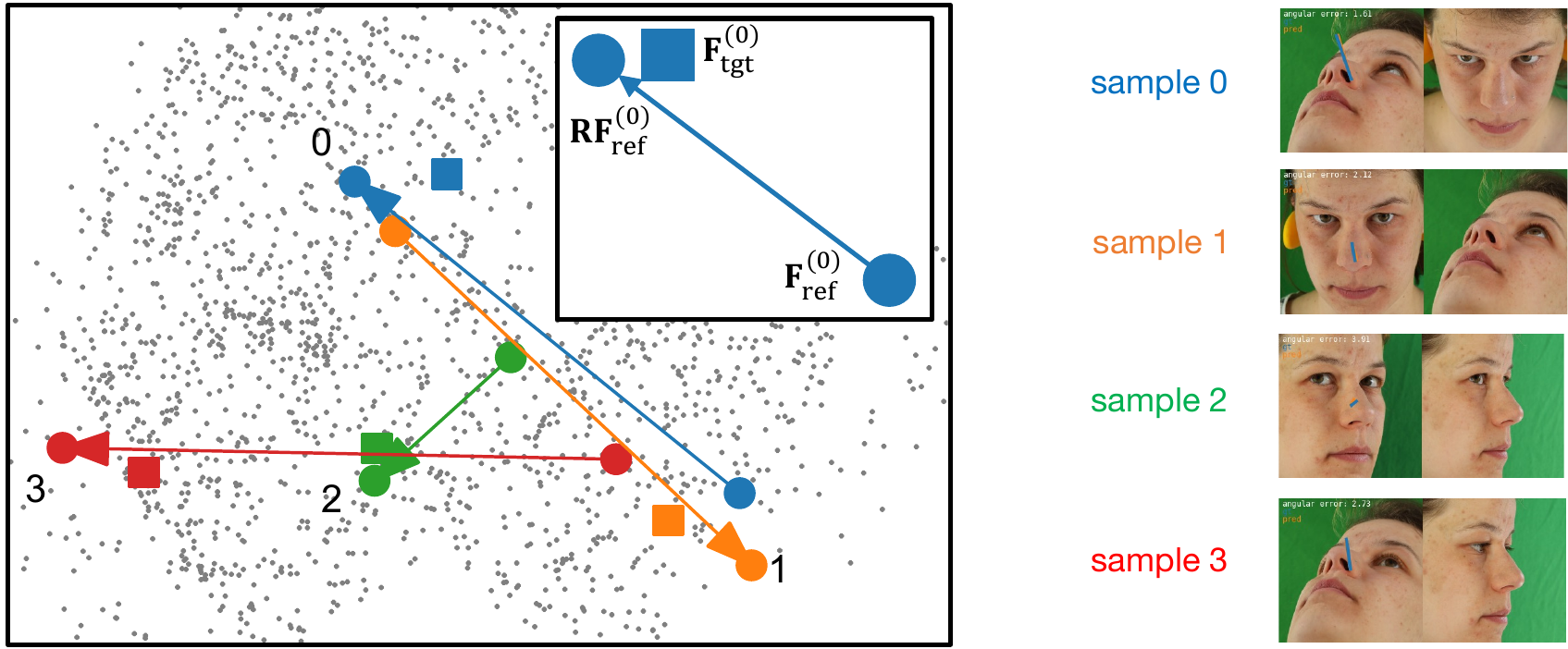}
    \caption{
            Isomap embedding of the initial rotatable features.
            The right side shows the example input samples.
            $\mathbf{F}^{(0)}_\mathrm{ref}$, $\mathbf{F}^{(0)}_\mathrm{tgt}$, and $\mathbf{RF}^{(0)}_\mathrm{ref}$ of the same sample are represented in the same color on the left side plot.
    }
    \label{fig:isomap-iter0}
\end{figure}

\Fref{fig:isomap-iter0} shows the features $\mathbf{F}^{(0)}_\mathrm{ref}$, $\mathbf{F}^{(0)}_\mathrm{tgt}$, and $\mathbf{RF}^{(0)}_\mathrm{ref}$ embedded in Isomap~\cite{isomap}.
We use the XGaze dataset under the unseen head pose condition.
Isomap embedding was generated from the initial rotatable features $\mathbf{F}^{(0)}$ obtained from 1000 test samples with a neighborhood size of $30$.
Marker shape indicates different feature types,
and color indicates the sample ID.
The arrows indicate the feature position before and after rotation.
Since the rotation is symmetric, we only visualize the rotation from the reference to the target.
We can clearly observe that the rotation operation brings the feature closer to the other.
This indicates that, as intended, the \emph{3D Feature Extractor} module learns to extract rotatable features through our rotation constraint.

\begin{figure}[t]
    \centering
    \includegraphics[width=0.9\linewidth]{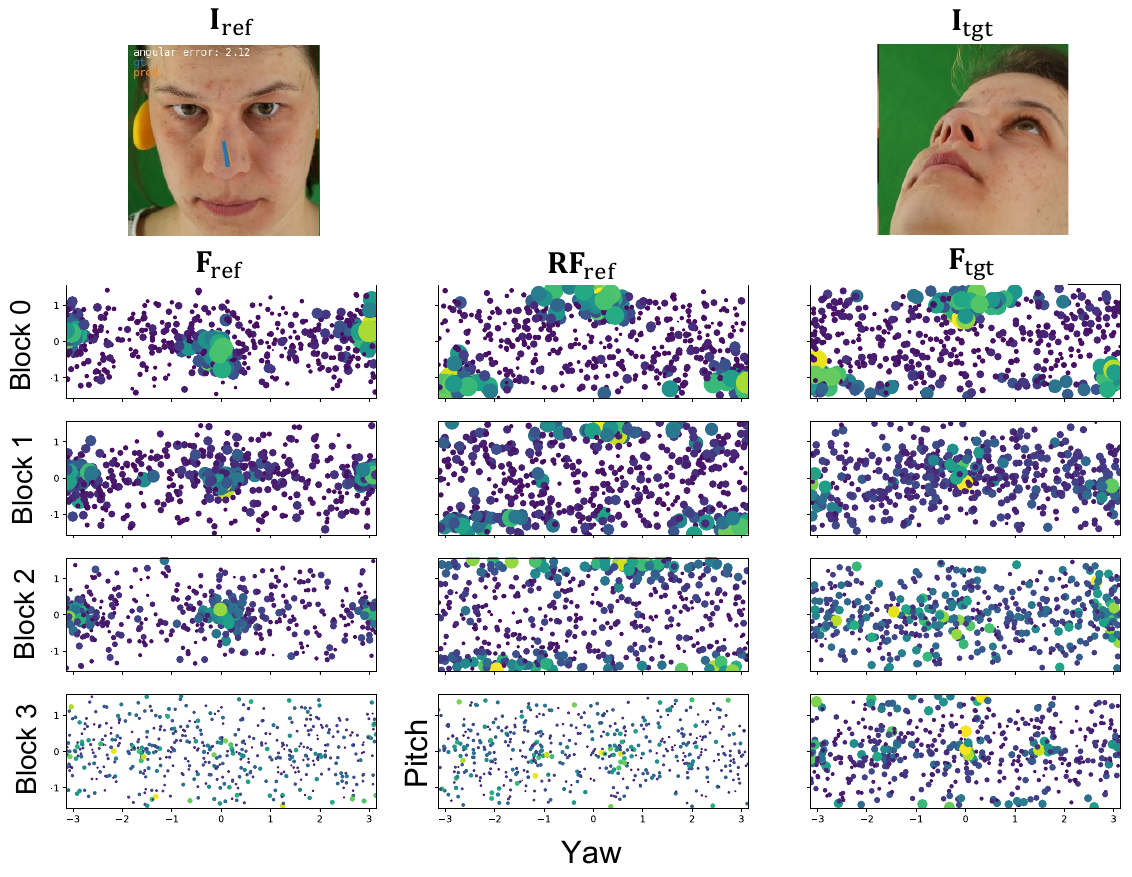}
    \caption{
        Scatter plot visualization of the rotatable features.
        Each of the $D$ 3D vectors is represented in a pitch-yaw coordinate system.
        Each row corresponds to the rotatable features at different fusion stages from $\mathbf{F}^{(0)}$ to $\mathbf{F}^{(3)}$.
        Larger and yellower dots represent elements with a larger norm.
    }
    \label{fig:feature-pitchyaws}
\end{figure}

\Fref{fig:feature-pitchyaws} shows an example of the rotatable features.
In this plot, we interpret the rotatable features as a set of $D$ 3D vectors and transform each 3D vector into the pitch-yaw coordinate system.
The size and color of the dots represent the magnitude of the norm of the 3D vectors, and large yellow dots indicate vectors with larger norms.
The distributions of $\mathbf{RF}^{(0)}_\mathrm{ref}$ and $\mathbf{F}^{(0)}_\mathrm{tgt}$ become closer by rotation before the first block, being consistent with \Fref{fig:isomap-iter0}.
Meanwhile, subsequent fusions with backbone features make the two features different in later blocks, \eg, $\mathbf{RF}^{(3)}_\mathrm{ref}$ and $\mathbf{F}^{(3)}_\mathrm{tgt}$.
We hypothesize that rotatable features adaptively evolve through stacked fusion blocks into complementary representations of the backbone features.

\subsection{Contribution of Reference Images}\label{sec:ref-contribution}
\begin{figure}[t]
    \centering
    \includegraphics[width=0.9\linewidth]{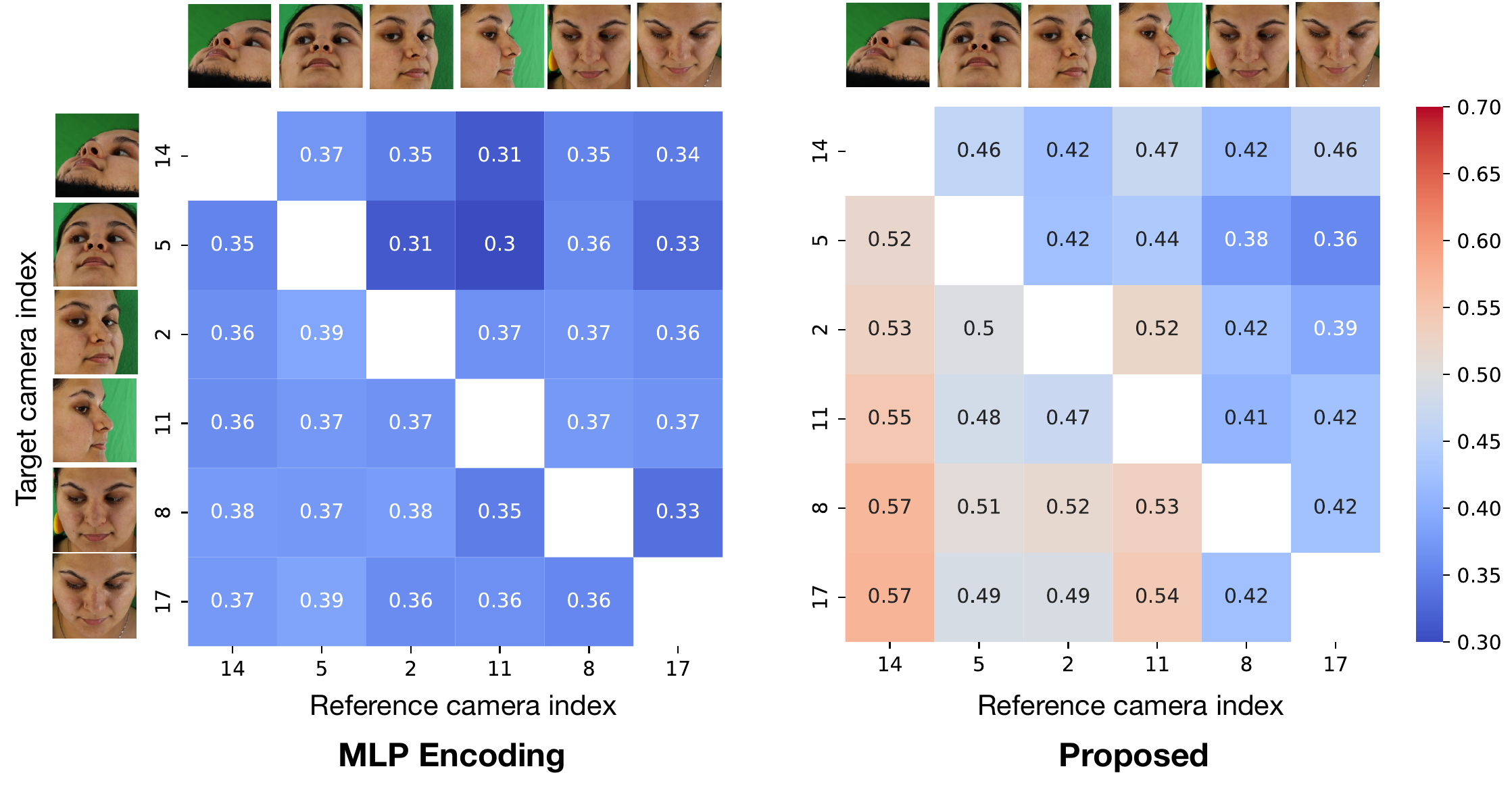}
    \caption{
        Analysis of contributions of features from each view.
        The values represent the contribution ratio of the feature from the reference images, which is calculated as the sum of the gradient of the backbone features.
    }
    \label{fig:contrib-xgaze}
\end{figure}

\Fref{fig:contrib-xgaze} illustrates the contribution ratio of the reference features for each camera pair.
As a metric for feature contribution, we calculated the sum of the gradient of the backbone features.
We use the XGaze dataset under the unseen head pose condition as the experimental setup.
A larger number represents more contribution of the reference image to the estimation result.
\Fref{fig:contrib-xgaze} shows the visualization results of the \emph{MLP Encoding} baseline (left) and our proposed method (right).
Comparing the two visualization results, we can confirm that our method adaptively uses the reference images.
While \emph{MLP Encoding} model always ignores the reference images, our method uses the reference information mainly depending on its head poses.

\begin{figure}[t]
    \centering
    \includegraphics[width=0.9\linewidth]{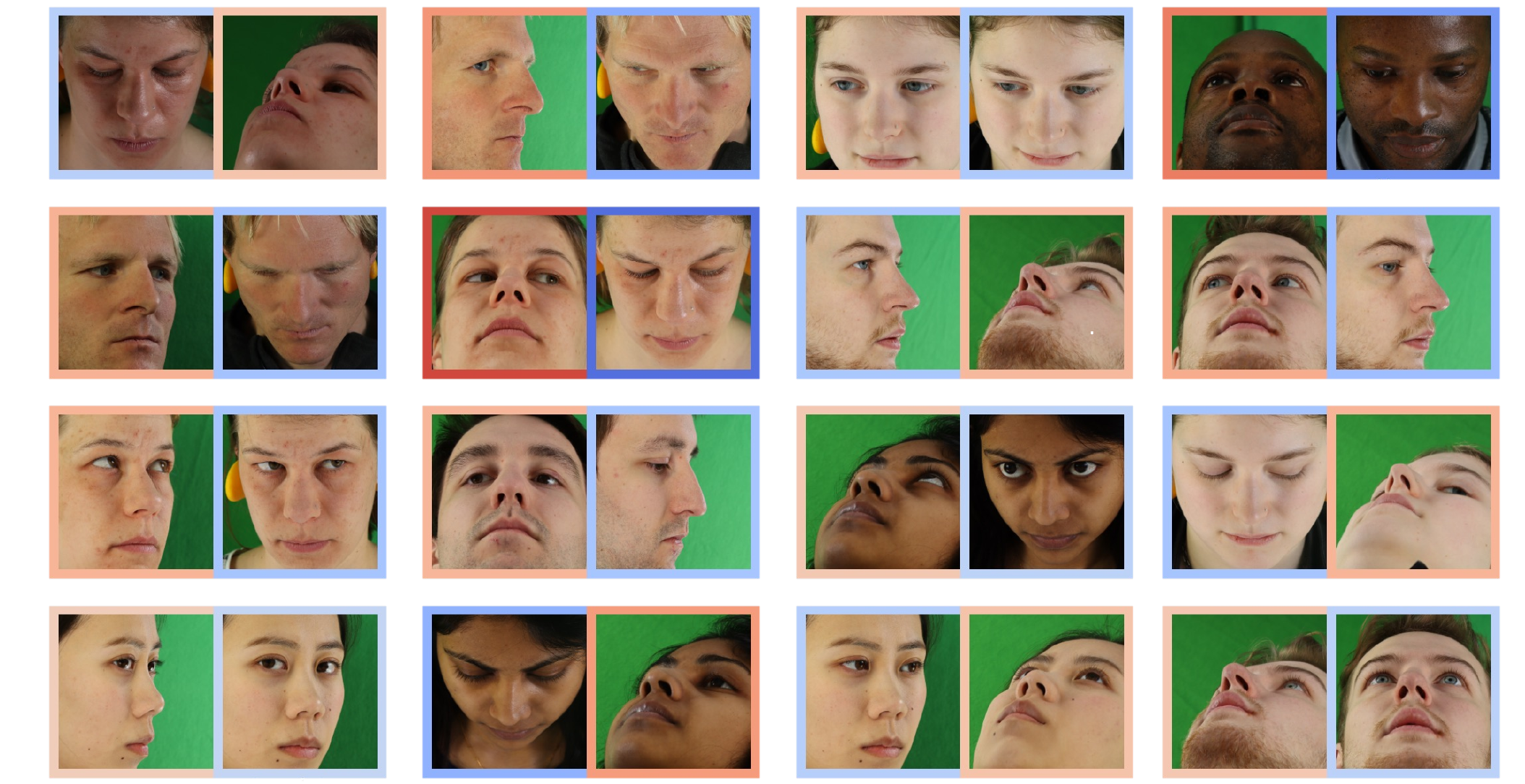}
    \caption{
        Example of paired images with their contribution to gaze estimation.
        The left and right images correspond to the target and the reference.
        The edge color shows the contribution ratio, where red indicates a higher contribution.
    }
    \label{fig:qual}
\end{figure}

\Fref{fig:qual} further shows sample images with their corresponding contribution ratios.
The edge color of each image represents its contribution as in \Fref{fig:contrib-xgaze}.
Overall, images with a view that captures the face from below have a higher contribution.
This is consistent with \Fref{fig:contrib-xgaze} where, \eg, camera 14 shows a more significant contribution.
Occlusion caused by eyelids is possibly a significant factor in the minor contribution of images captured from the top view.

\section{Conclusion}
\label{cha:conclusion}

In this paper, we presented a novel multi-view appearance-based gaze estimation task.
We propose a cross-view feature fusion approach using the relative rotation matrix between input images as a constraint when transferring the features to the other image.
In addition to its practical significance, the proposed method has the advantage of improving generalization performance for unseen head poses.
Through experiments, we demonstrated the advantage of our method over state-of-the-art baseline, including single-image domain generalization methods.

The limitation of our approach compared to a single-image baseline is the slightly increased hardware requirements.
The requirements of our method are not particularly unrealistic compared to existing eye trackers, and this is ultimately a matter of trade-offs.
The same can be said about the effect of camera calibration.
It is also essential for future work to develop lightweight models that are robust to errors in the rotation matrix and time synchronization.

\subsection*{Acknowledgement}
This work was supported by JSPS KAKENHI Grant Number JP21K11932.

{\small
\bibliographystyle{ieee_fullname}
\bibliography{egbib}
}

\clearpage
\setcounter{page}{1}
\setcounter{figure}{0}
\setcounter{table}{0}
\setcounter{equation}{0}

\appendix
\section{Detailed Ablation Studies}

\begin{table}[t]
    \centering
    \begin{tabular}{@{}lccc@{}}
        \toprule
         Head pose & Seen & Unseen \\ \midrule
         w/o Separate Fusers & 3.49\textdegree & 5.10\textdegree \\
         w/o Backbone Features & \textbf{3.46\textdegree} & 5.20\textdegree \\
         Proposed & 3.50\textdegree & \textbf{4.95\textdegree} \\ \bottomrule
    \end{tabular}
    \caption{
        Ablation studies of learnable modules.
        We ablated the separate weight of the \textit{Fusers} and \textit{3D Feature Extractor}.
        }
    \label{tab:ablation-extractor}
\end{table}

We perform ablation studies on several learnable modules of the proposed method to validate our design choice on \emph{Fuser}.
The first row (\emph{w/o Separate Fusers}) in \Tref{tab:ablation-extractor} corresponds to a variant of the proposed method where the \emph{Fusers} in each block share the same weights.
The model in the second row (\emph{w/o Backbone Features}) uses the initial rotatable feature $\mathbf{F}^{(0)}$ as input to \emph{Fusers} instead of the backbone feature $\mathbf{f}$.
This model, therefore, does not distinguish between rotatable and backbone features.

While both methods perform on par with the proposed method under the \emph{seen} setting, the proposed method shows superiority in the \emph{unseen} setting.
One possible explanation is that stacking different fusion blocks allows the model to focus on different patterns depending on the depth of the block and that the original backbone feature still contains valuable information for appearance-based gaze estimation.

\section{Visualization of Rotatable Features}

\begin{figure}[t]
    \centering
    \includegraphics[width=\linewidth]{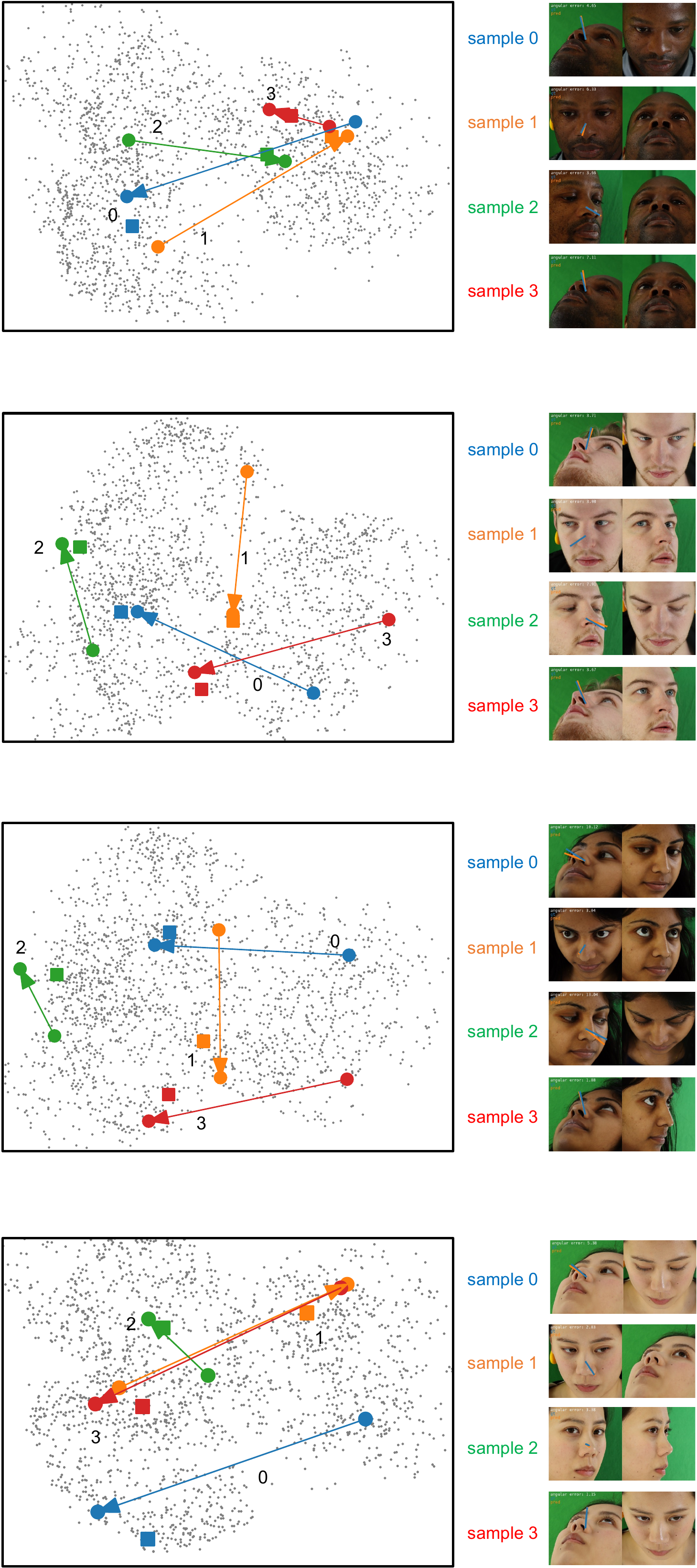}
    \caption{
            Isomap embedding of the initial rotatable features.
            The right side shows the example input samples.
            $\mathbf{F}^{(0)}_\mathrm{ref}$, $\mathbf{F}^{(0)}_\mathrm{tgt}$, and $\mathbf{RF}^{(0)}_\mathrm{ref}$ of the same sample are represented in the same color on the left side plot.
    }
    \label{fig:isomap-samples}
\end{figure}

In \Fref{fig:isomap-samples}, we depict more Isomap embedding of the initial rotatable features from test subjects.
Each Isomap embedding was generated from the features obtained from each target participant, and all other visualization details are consistent with the main paper.
The visualization results confirm that the proposed method acquires person-independent rotatable feature representations.

\begin{figure*}[ht]
    \centering
    \includegraphics[width=\linewidth]{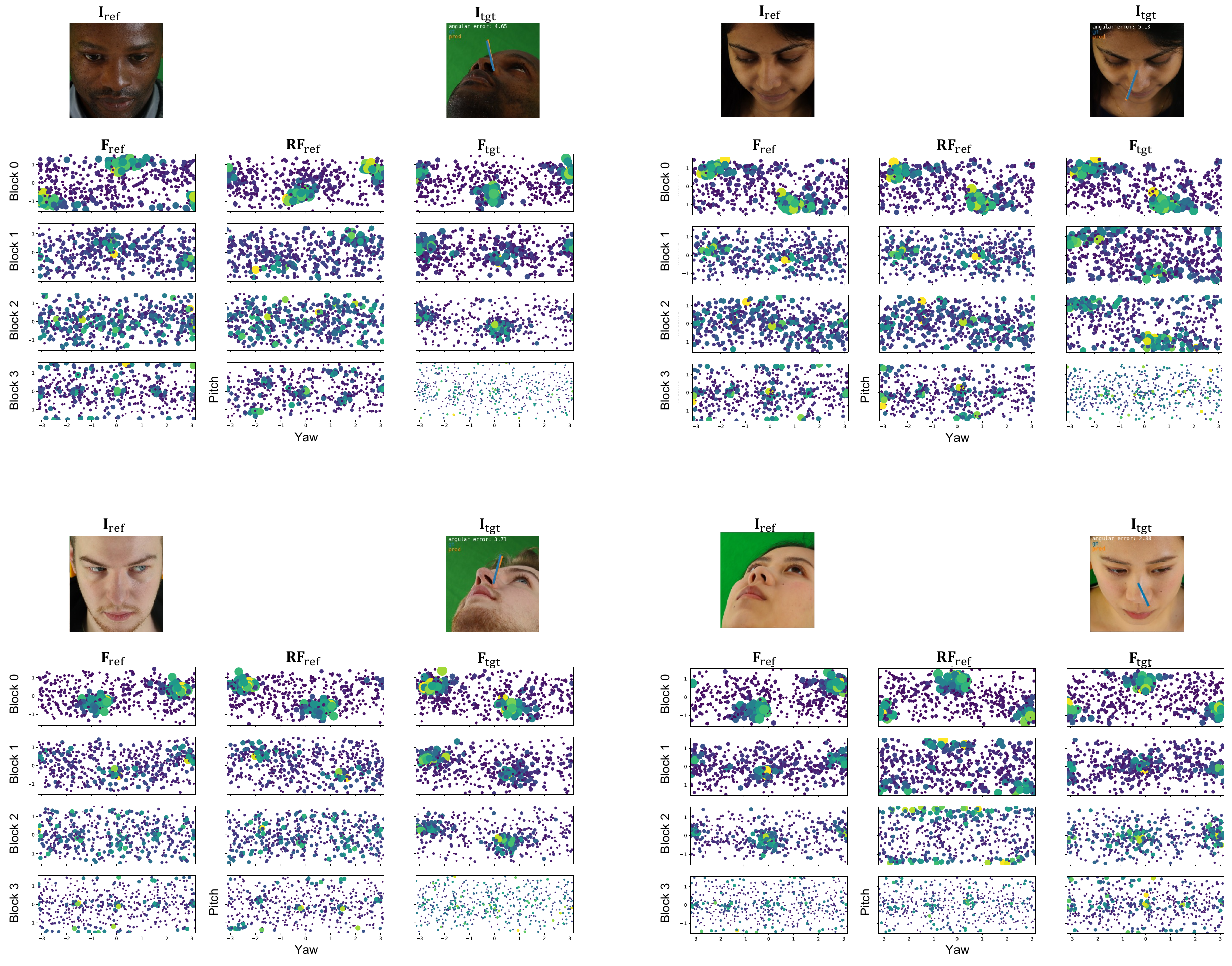}
    \caption{
        Scatter plot visualization of the rotatable features.
        Each of the $D$ 3D vectors is represented in a pitch-yaw coordinate system.
        Each row corresponds to the rotatable features at different fusion stages.
        Larger and yellower dots represent elements with a larger norm.
    }
    \label{fig:scatter-samples}
\end{figure*}

In \Fref{fig:scatter-samples}, we also show more scatter plot visualizations of the rotatable features from test subjects in the yaw-pitch coordinate system. 
We can consistently observe the tendency for feature distributions to converge before the first fusion block and then diverge in later blocks across different subjects.
It can be seen that the proposed method dynamically updates rotatable features even with a slight rotation (the upper right example in \Fref{fig:scatter-samples}).

\section{Baseline Implementation Details}

Unless otherwise noted, all baseline methods follow the same training hyperparameters as used for the proposed method in the main paper.
We note that we did not tune the hyperparameters in favor of the proposed method. Instead, we used common choices, most of which already comply
with ResNet and PureGaze. With the Cyclic LR scheduler, ResNet, PureGaze, and Hybrid-TR are less tuning demanding.
Therefore we tune the training-unstable DT-ED.

\paragraph{DT-ED}
Since we use a richer full-face patch instead of an eye-region patch as input of DT-ED, we modified the appearance and gaze latent code sizes from 64 and 2 to 512 and 16.
Following the original setting, we used angular loss for gaze estimation and $\ell_1$ for reconstruction.
For the learning rate, we found that the scaling and ramp-up settings in the original paper make it difficult for the model to reconstruct the target image.
Therefore, we trained the model with a base learning rate of $5 \times 10^{-4}$ decaying by $0.8$ every 1 epoch, similar to another gaze redirection work~\cite{sted}. 
Unlike other baselines, the batch size is set to 60.

\paragraph{Gaze-TR}
In our implementation, we used ResNet-50~\cite{resnet} to extract feature maps from the images.
The size of the feature map was $7 \times 7 \times 32$, which is then fed to a six-layer transformer.
Finally, an MLP takes the feature vector as input and estimates the gaze direction.

\paragraph{PureGaze}
When training models on both ETH-XGaze~\cite{xgaze} and MPII-NV~\cite{jiawei} dataset, we used the default mask image in the official PureGaze repository~\footnote{https://github.com/yihuacheng/PureGaze} generated for normalized ETH-XGaze face images to compute the adversarial reconstruction loss. 
For the extra hyperparameters controlling the relative contribution of the adversarial loss to the total loss, we followed the official implementation.

\section{Definition of the Rotation Matrix}

As discussed in the paper, there are two approaches to computing the relative rotation matrix $\mathbf{R}$ using either camera calibration or head poses estimation.
In the following, we provide detailed explanations of two claims: 1) the final $\mathbf{R}$ becomes the same in either approach, and 2) the relative translation $\mathbf{t}$ is uniquely determined by $\mathbf{R}$ and can be ignored.

First, we show that the two definitions $\mathbf{R} = \mathbf{N}_{\mathrm{tgt}} \tilde{\mathbf{R}} \mathbf{N}^{\top}_{\mathrm{ref}}$ and $\mathbf{R} = \mathbf{H}_\mathrm{tgt} \mathbf{H}^{\top}_\mathrm{ref}$ are interconvertible.
Let us denote the camera extrinsic parameters, \ie, the transformation from the reference to the target camera coordinate systems, as $\mathbf{C} \in \mathbb{R}^{4 \times 4}$.
If we denote head poses in the original camera coordinate systems before normalization as $\hat{\mathbf{H}} \in \mathbb{R}^{4 \times 4}$, their relationship can be defined as 
\begin{equation}
\hat{\mathbf{H}}_\mathrm{tgt} = \mathbf{C} \hat{\mathbf{H}}_\mathrm{ref}. \label{eq:extrinsic}
\end{equation}
If we further denote an extended $4 \times 4$ normalization matrix as $\mathbf{N}$, the head poses $\mathbf{H}$ after normalization can also be obtained from the normalization matrix as 
\begin{equation}
\mathbf{H} = \mathbf{N} \hat{\mathbf{H}}. \label{eq:normalize}
\end{equation}
From \Eref{eq:extrinsic} and \Eref{eq:normalize}, we can derive that
\begin{eqnarray*}
    \mathbf{N}_\mathrm{tgt} \mathbf{C} \mathbf{N}^{\top}_\mathrm{ref} &=&
    \mathbf{N}_\mathrm{tgt} \hat{\mathbf{H}}_\mathrm{tgt} \hat{\mathbf{H}}^{\top}_\mathrm{ref} \mathbf{N}^{\top}_\mathrm{ref} \label{eq:def} \\
    &=&
    (\mathbf{N}_\mathrm{tgt} \hat{\mathbf{H}}_\mathrm{tgt}) (\mathbf{N}_\mathrm{ref} \hat{\mathbf{H}}_\mathrm{ref})^{\top} \\
    &=& \mathbf{H}_\mathrm{tgt} \mathbf{H}^{\top}_\mathrm{ref}.
\end{eqnarray*}
Therefore, we can conclude that the two definitions are interconvertible and have the same meaning.
Note that this applies not only to the rotation component $\mathbf{R}$ but also to the translation component $\mathbf{t}$.

Next, we show that the translation component $\mathbf{t}$ is uniquely determined by the rotation $\mathbf{R}$ under the assumption of data normalization.
One of the key properties of the normalization process is that the origin of the gaze vector is located at a fixed distance $d$ on the $z$-axis of the camera coordinate system.
Therefore, this origin $\mathbf{o} = (0, 0, d, 1)^{\top}$ does not move when the above transformation matrix is applied:
\begin{eqnarray} \label{eq:cond}
    \mathbf{o}_\mathrm{tgt}
    = \begin{pmatrix}
        \mathbf{R} & \mathbf{t} \\
        0 & 1
    \end{pmatrix}
    \mathbf{o}_\mathrm{ref} = \mathbf{o}_\mathrm{ref}.
\end{eqnarray}

If we denote $\mathbf{R} = (\mathbf{r}_x, \mathbf{r}_y, \mathbf{r}_z)$ where $\mathbf{r}_x, \mathbf{r}_y, \mathbf{r}_z \in \mathbb{R}^3$ are the column vectors of the rotation matrix, substituting this into \Eref{eq:cond} yields
\begin{eqnarray*}
    \begin{pmatrix}
        0 \\ 0 \\ d \\ 1
    \end{pmatrix} &=& 
    \begin{pmatrix}
        \mathbf{r}_x & \mathbf{r}_y & \mathbf{r}_z & \mathbf{t} \\
        & 0 & & 1
    \end{pmatrix}
    \begin{pmatrix}
        0 \\ 0 \\ d \\ 1
    \end{pmatrix} \\
    &=&
    \begin{pmatrix}
        d \mathbf{r}_z + \mathbf{t} \\ 1
    \end{pmatrix}.
\end{eqnarray*}
Therefore, the translation component $\mathbf{t}$ is uniquely defined by the fixed distance $d$ and the rotation vector $\mathbf{r}_z$ as
\begin{equation}
    \mathbf{t} =
    \begin{pmatrix}
        0 \\ 0 \\ d
    \end{pmatrix}
     - d \mathbf{r}_z,
\end{equation}
and can be ignored in our problem setting.


\end{document}